\documentclass[conference]{IEEEtran}
\usepackage[left=1in,right=1in,top=0.75in,bottom=1in]{geometry}

\usepackage{cite}
\usepackage{amsmath,amssymb,amsfonts}
\usepackage{colortbl}
\usepackage{graphicx}
\usepackage{textcomp}
\usepackage{subfig}
\usepackage{balance}
\usepackage{float}	
\usepackage{soul}
\usepackage{xcolor}
\usepackage{listings}   
\usepackage{hyperref}
\def\BibTeX{{\rm B\kern-.05em{\sc i\kern-.025em b}\kern-.08em
T\kern-.1667em\lower.7ex\hbox{E}\kern-.125emX}}

\begin{document}

\title{An Application of Vector Autoregressive Model for Analyzing the Impact of Weather And Nearby Traffic Flow On The Traffic Volume}

\author{\IEEEauthorblockN{Anh Thi-Hoang Nguyen\IEEEauthorrefmark{1}$^,$\IEEEauthorrefmark{2},  Dung Ha Nguyen\IEEEauthorrefmark{1}$^,$\IEEEauthorrefmark{2}, Trong-Hop~Do\IEEEauthorrefmark{1}$^,$\IEEEauthorrefmark{2}}\vspace{2ex}
        \IEEEauthorblockA{\IEEEauthorrefmark{1} University of Information Technology, Ho Chi Minh City, Vietnam.\\
        \IEEEauthorrefmark{2} Vietnam National University, Ho Chi Minh City, Vietnam.
        }
}

\maketitle

\begin{abstract}
This paper aims to predict the traffic flow at one road segment based on nearby traffic volume and weather conditions. Our team also discover the impact of weather conditions and nearby traffic volume on the traffic flow at a target point. The analysis results will help solve the problem of traffic flow prediction and develop an optimal transport network with efficient traffic movement and minimal traffic congestion. Hourly historical weather and traffic flow data are selected to solve this problem. This paper uses model VAR(36) with time trend and constant to train the dataset and forecast. With an RMSE of 565.0768111 on average, the model is considered appropriate although some statistical tests implies that the residuals are unstable and non-normal. Also, this paper points out some variables that are not useful in forecasting, which helps simplify the data-collecting process when building the forecasting system.

\begin{IEEEkeywords}Traffic Flow Analysis, Historical Weather, Time Series, VectorAutoRegression, VAR, Multivariate, Impact analysis
\end{IEEEkeywords}
\end{abstract}

\setcounter{secnumdepth}{4}

\section{Introduction}
Transportation is one of the most influential aspects in every country around the world. Each day, thousands of people rush onto the roads and travel by different means of transportation. A high volume of transport or traffic jams may considerably influence people's prearranged plans. More negatively, accidents and even deaths could also occur. Therefore, there is a practical need for an effective methodology of traffic flow prediction to avoid unexpected events. Accurately forecasting traffic flow in real-time could improve the operational efficiency of the city and support information management decision-making. Additionally, accurate prediction results could optimize each citizen's travel planning and save their travel time.

Identifying the characteristics of the traffic network has always been a challenging issue. Numerous works have been done on this topic using different approaches, including neural networks, regression models, and univariate statistic models such as AR (Autoregressive), MA (moving average), ARMA (Autoregressive Moving Average), and ARIMA (Autoregressive Integrated Moving Average). Although most of the research obtained positive results, there exist several limitations. The most common approach has been univariate time series models, while multivariate one enables researchers to explore more about the impacts of other variables on the object. This paper assumes that traffic flow at a specific time depends not only on previous data of its time series but also on the traffic state at surrounding places and weather conditions. Therefore, there is a likelihood that adding traffic and weather attributes can help predict traffic volume at a target road section. Based on the assumption, this paper will apply a multivariate time series model called Vector Autoregressive Model (VAR) to do the forecast. The choice of research method is adjusted to the research data, which has the characteristic of a stationary time series. 
In this study, we collected traffic and weather data recorded in Dublin, Ireland from March 01, 2013 to February 28, 2019. Then, we perform several analysis to clarify if the information about nearby traffic state and various weather conditions helpful for forecasting the traffic flow at the target. Then we conclude which weather data to be unnecessary for predicting traffic flow. This finding helps simplify the data collecting process when building a traffic flow forecasting system.

\section{Methodology}
\begin{figure*}[!h]
    \centering
    \includegraphics[width = 16 cm]{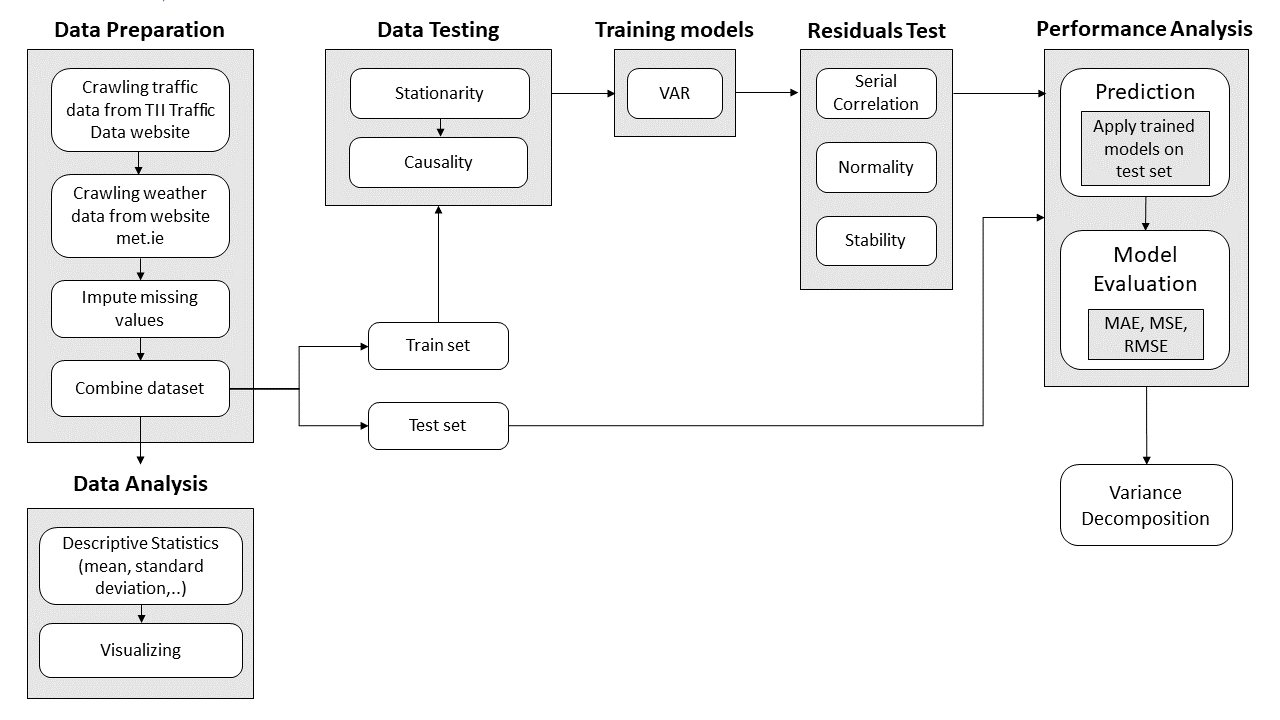}
    \caption{The analytical flow of this study}
    \label{fig:analyticalflow}
\end{figure*}

The analytical flow is shown in Fig. \ref{fig:analyticalflow}. First, we crawl traffic and weather data. Then we preprocess and combine these two dataset to make the dataset which is ready for analysis. The combined dataset is splited into train and test dataset. The train dataset is used for examine time series features such as stationarity and causality. After this examining step, the train dataset is used to train VAR model. The prediction results are then applied residuals test. After that, the trained model is applied to the test dataset and the performance of the model is evaluated through Mean Absolute Error (MAE), Mean Squared Error (MSE), and Root Mean Square Error (RMSE). Finally, we apply variance decomposition to show how much information each variable adds to other variables. 
\subsection{Stationarity}
It is typical to examine the stationarity of all the time series in the system because the time series you want to predict must be steady in order to use the VAR model. If the statistical characteristic remains constant over time, a time series is said to be stationary. By differencing, a non-stationary series $X_{t}$ can be made stationary:
\begin{equation}
    x_{t}=X_{t}-X_{t-1}
    \label{eq:stationarity}
\end{equation}
\\It can be concluded that stationary data will tend to approach its average value and fluctuate around its mean value. More specifically, this study will use the Augmented Dickey-Fuller test (ADF), which is a common statistical test used to test whether a given Time series is stationary or not. When examining the stationary of a series, it is one of the statistical tests that is most frequently employed.
\begin{equation}
 \begin{array}{l}
y_{t} =c+\beta t+\alpha y_{t-1} +\phi _{1} \Delta Y_{t-1} +\phi _{2} \Delta Y_{t-2} +...\\
+\phi _{p} \Delta Y_{t-p} +e_{t}
\end{array}
\label{eq: adftestformula}
\end{equation}

\subsection{Granger Causality}
The Granger causality approach was used to study the causal link structures between variables. A statistical test to see if one time series can accurately predict another is called the Granger causality test. The hypothesis would be rejected at that level if the probability value is less than any alpha level. To put it simply, time series (X) previous values do not influence the other series (Y).
\\\\Therefore, we can safely reject the null hypothesis if the p-value returned from the test is lower than the significance level of 0.05. The Granger Causality formula is shown below:
\begin{equation}
    X_{1}( t) =\sum _{j=1}^{p} A_{11,j} X_{1}( t-j) +\sum\limits _{j=1}^{p} A_{12,j} X_{2}( t-j) +E_{1}( t)
    \label{eq:grangercausality1}
\end{equation}
\begin{equation}
    X_{2}( t) =\sum _{j=1}^{p} A_{21,j} X_{1}( t-j) +\sum _{j=1}^{p} A_{22,j} X_{2}( t-j) +E_{2}( t)
    \label{eq:grangercausility2}
\end{equation}
\\where p is the model order, A is the matrix of coefficients, and $E_{1}$ and $E_{2}$ are prediction errors (residuals) for each time series.

\subsection{Vector Autoregressive Model (VAR)}
A statistical technique called vector autoregressive is used to examine the correlation between a number of influencing factors. Because they are adaptable and straightforward models for multivariate time series data, vector autoregressive (VAR) processes are well-liked in the fields of economics and other sciences. It can be applied to situations where several factors are interdependent. Each variable in a VAR model is modeled as a linear mixture of historical data for both it and other variables.

As a result, it can be modeled as a set of equations, where each variable is given a set of vectorial equations. Let's say we have a time series data vector $Y_{t}$, then a VAR model where $Y_{t}$, $\beta_{0}$ and are k x 1 column vectors, and $\beta_{0}$, $\beta_{1}$,... $\beta_{p}$ are k x k matrices of coefficients with p lags can be expressed mathematically in equation (5):
\begin{equation}
   y_{t} =\beta _{0} +\beta _{1} \Delta Y_{t-1} +\beta _{2} \Delta Y_{t-2} ..+\beta _{p} \Delta Y_{t-p} +e_{t}
       \label{eq:varmodel}
\end{equation}
\\\\If a time series is not stationary, it is essential to differentiate the time series before training the model and invert the predicted values to get the actual forecast by the number of times differentiated. 

\subsection{Testing for Residuals}
\subsubsection{Check for Serial Correlation}
To determine if the residuals still contain any patterns, serial correlation of the residuals (errors) is performed. It has long been difficult to test for serial correlation in time series. The Durbin-Watson test, the first formally developed method for testing first-order serial correlation, is the most well-known test for serial correlation in regression disturbances. Durbin and Watson (1950,1951) specifically suggest the following test statistic:

\begin{equation}
    d=\frac{\sum _{t=2}^{n}( e_{t} -e_{t-1})^{2}}{\sum _{t=2}^{n} e_{t}^{2}}
        \label{eq:durinwatsontest}
\end{equation}
\subsubsection{Check for Normality}
A model must do a diagnostic analysis to identify before it can be utilized for predicting. A test to ascertain the normalcy residual of data is testing for normality of residual. This test's goal is to determine whether or not the residuals from the data are distributed regularly. We can utilize the Jarque-Bera (JB) Test of Normality to determine whether something is normal. Skewness and kurtosis measurements were employed in this test. The calculation of JB is as follows:
\begin{equation}
    JB=\frac{n}{6}\left( s^{2} +\frac{( k-3)^{2}}{4}\right) =\frac{\frac{1}{n}\sum _{i=1}^{n}( X_{i} -\overline{X})^{4}}{\left(\frac{1}{n}\sum _{i=1}^{n}( X_{i} -\overline{X})^{2}\right)^{2}}
        \label{eq:jarqueberatest}
\end{equation}
\\where n is the number of samples, s is expected Skewness and k is expected excess kurtosis
\subsubsection{Check for Stability}
If a process's reverse characteristic polynomial of the VAR(p) has no roots in or on the complex unit circle, it is said to be stable. Equation is spelled out as follows:
\begin{equation}
    y_{t} =c+\phi _{1} \Delta Y_{t-1} +\phi _{2} \Delta Y_{t-2} ..+\phi _{p} \Delta Y_{t-p} +e_{t}
        \label{eq:stability}
\end{equation}

\subsection{Variance Decomposition}
Once a vector autoregression (VAR) model has been fitted, it can be interpreted using a variance decomposition or forecast error variance decomposition (FEVD), which is used in econometrics and other applications of multivariate time series analysis. The variance decomposition shows how much information each variable in the autoregression adds to the other variables. It establishes the percentage of each variable's forecast error variation that exogenous shocks to the other variables may account for. The orthogonalized impulse responses $\Theta _{i}$ can be used to decompose the forecast errors of component j on k in an i-step forward forecast.
\begin{equation}
    w_{jk,i} =\sum _{i=0}^{h-1}\left( e_{j}^{'} \Theta _{i} e_{k}\right)^{2} /MSE_{j}( h)
        \label{eq: variancedecomposition1}
\end{equation}
\begin{equation}
    MSE_{j}( h) =\sum _{i=0}^{h-1} e_{j}^{'} \Phi _{i} \Sigma _{u} \Phi _{i}^{'} e_{j}
 \label{eq: variancedecomposition2}
\end{equation}

\subsection{Evaluation Metrics}
In time series forecasting,  to evaluate the models,  a comprehensive evaluation criterion is essential to measure the performance of the model. Root Mean Squared Error (RMSE), Mean Absolute Error (MAE), and Mean Squared Error (MSE) are commonly used metrics to reliably evaluate the performance of the models.
\begin{equation}
    RMSE=\sqrt{\sum\limits _{i=1}^{n}\frac{(\hat{y}_{i} -y_{i})^{2}}{n}}
    \label{eq:rmse}
\end{equation}
\begin{equation}
    MAE=\sum\limits _{i=1}^{n}\frac{| y_{i} -x_{i}| }{n}
    \label{eq:mae}
\end{equation}
\begin{equation}
    MSE=\frac{\sum ( y_{i} -\hat{y}_{i})^{2}}{n}
    \label{eq:mse}
\end{equation}

\section{Data Description and Analysis}
\subsection{Traffic Dataset}
\begin{figure}[!h]
    \centering
    \includegraphics[width= 8cm]{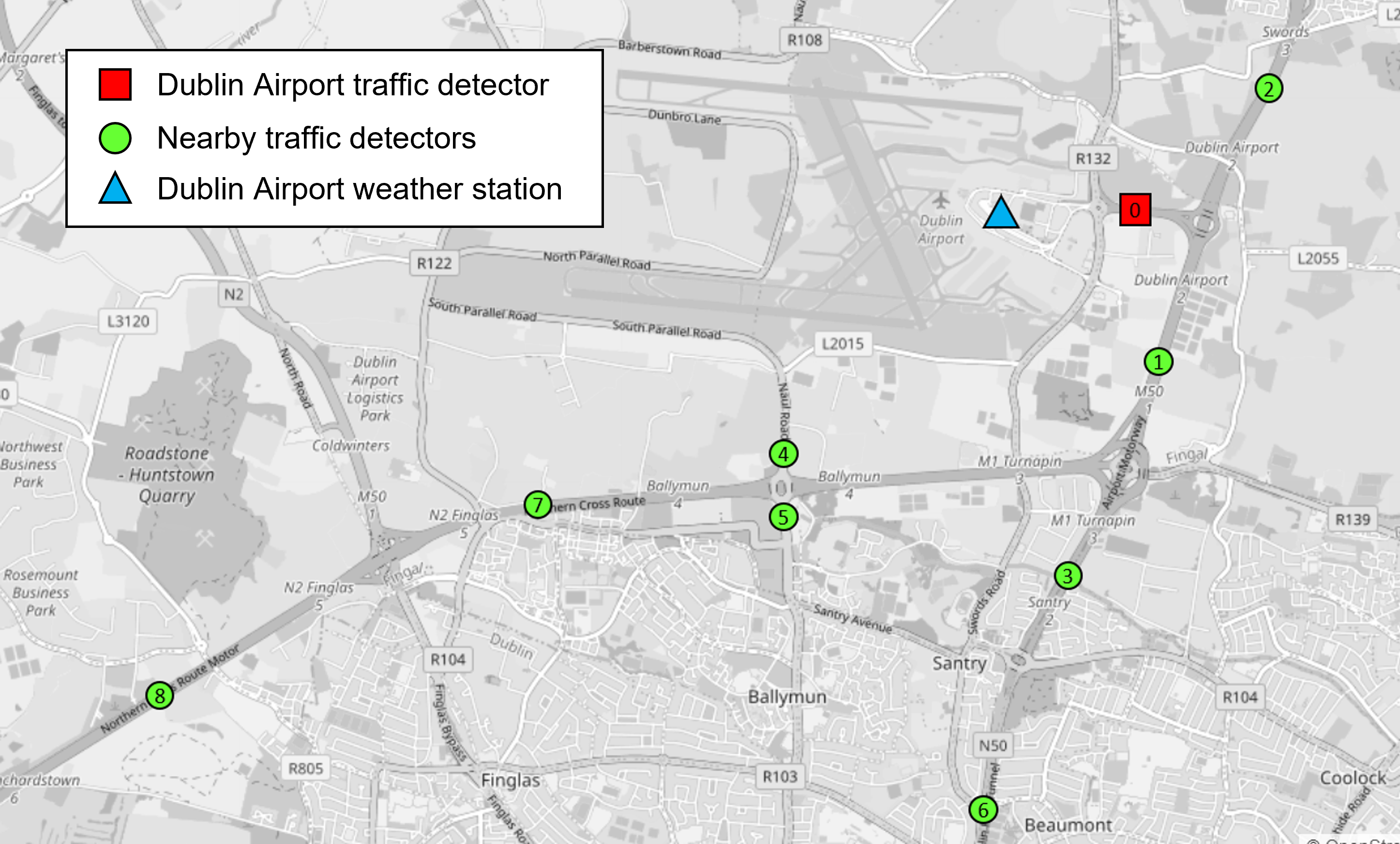}
    \caption{ Traffic counters and weather station distribution}
    \label{fig:trafficandweathercounters}
\end{figure}

This paper takes the traffic information collected by Transport Infrastructure Ireland (TII) traffic counters located on the road network of Ireland as the basis for analysis. The section where the counter of Dublin Airport link road (Station ID TMU N01 000.0 N) located is considered the target section. All counters within the radius of 10km, taking the target section as the center, are selected. There are 11 alternative counter sections found, among which three counters have a considerable amount of missing values during data extraction. Therefore, three counters containing insufficient data will be excluded from the dataset used for this study. The specific location and information of the counters are shown in Figure \ref{fig:trafficandweathercounters}.

Figure \ref{fig:trafficandweathercounters} shows the node locations of 8 independent traffic counters. All counters sample the traffic information of the section every hour from March 01, 2013 to February 28, 2019. The data generated by the traffic counter mainly includes traffic flow by direction and vehicles. This paper will use the total traffic flow for research. However, raw data collected from the website still contains some missing values. Several missing value imputations are consistently implemented to enhance the analysis performance.

\subsection{Weather Dataset}
Dublin hourly historical weather data for the years surveyed is acquired from the website Met Éireann, Ireland’s National Meteorological Service. With the intention of accurately inspecting the interrelation between traffic flow and weather conditions, data at the nearest station from the Dublin Airport traffic detector are collected. The location of the Dublin Airport weather station is illustrated in Figure \ref{fig:trafficandweathercounters}.
\\Based on the purpose of this study, we drop out unnecessary weather attributes, and only important ones are selected, as shown in Table \ref{tab: weatherdecription}. This dataset does not contain any missing values; hence no imputation is needed.
\begin{table}[!h]
\centering
\begin{tabular}{|ccc|}
\hline
\textbf{Variable} & \textbf{Meaning}           & \textbf{Unit}                 \\ \hline
rain              & Precipitation Amount       & mm                            \\
temp              & Air Temperature            & °C                            \\
wetb              & Wet Bulb Temperature       & °C                            \\
dewpt             & Dew Point Temperature      & °C                            \\
rhum              & Relative Humidity          & \%                            \\
vappr             & Vapour Pressure            & hPa                           \\
msl               & Mean Sea Level Pressure    & hPa                           \\
wdsp              & Mean Wind Speed            & knot                          \\
wddir             & Predominant Wind Direction & degree                        \\
sun               & Sunshine duration          & hours                         \\
vis               & Visibility                 & m                             \\
vlht              & Cloud height               & ft  \\
clamt             & Cloud amount               & -                             \\ \hline
\end{tabular}
\caption{Weather Attributes Description}
\label{tab: weatherdecription}
\end{table}

\section{Results and Discussion}
\subsection{Stationary check using ADF test}
Hypothesis:

H0 : Data has unit root, hence not stationary

H1 : Data has no unit root, hence stationary
\\\\Table \ref{tab: adftest} indicates the results of the ADF unit root test. At the significant level of 0.05, ADF test statistics of all variables are less than critical values 5\%, with p-value of 0.0 less than the significant level. Thus, null hypothesis related to the unit root test is rejected; and the series are taken to be stationary. After this, Granger causality analysis can proceed among variables.\\
\\Critical Value 1\%: -3.43
\\Critical Value 5\%: -2.862
\\Critical Value 10\%: -2.567

\begin{table}[!h]
\centering
\resizebox{\columnwidth}{!}{%
\begin{tabular}{
>{\columncolor[HTML]{F3F3F3}}c cc
>{\columncolor[HTML]{F3F3F3}}c cc}
\hline
\textbf{Variable} & \textbf{\begin{tabular}[c]{@{}c@{}}ADF Test\\ Statistic\end{tabular}} & \textbf{P-Value} & \textbf{Variable} & \textbf{\begin{tabular}[c]{@{}c@{}}ADF Test\\ Statistic\end{tabular}} & \textbf{P-Value} \\ \hline
\textbf{station0} & -14.5967                                                              & 0.0              & \textbf{wetb}     & -10.0818                                                              & 0.0              \\
\textbf{station1} & -11.7656                                                              & 0.0              & \textbf{dewpt}    & -10.7832                                                              & 0.0              \\
\textbf{station2} & -26.8873                                                              & 0.0              & \textbf{rhum}     & -10.7947                                                              & 0.0              \\
\textbf{station3} & -22.7503                                                              & 0.0              & \textbf{vappr}    & -20.0823                                                              & 0.0              \\
\textbf{station4} & -22.2427                                                              & 0.0              & \textbf{msl}      & -15.0353                                                              & 0.0              \\
\textbf{station5} & -29.7172                                                              & 0.0              & \textbf{wdsp}     & -20.9809                                                              & 0.0              \\
\textbf{station6} & -19.3717                                                              & 0.0              & \textbf{wddir}    & -17.4869                                                              & 0.0              \\
\textbf{station7} & -24.5004                                                              & 0.0              & \textbf{sun}      & -21.8096                                                              & 0.0              \\
\textbf{station8} & -25.044                                                               & 0.0              & \textbf{vis}      & -21.0208                                                              & 0.0              \\
\textbf{rain}     & -25.6874                                                              & 0.0              & \textbf{clht}     & -27.1395                                                              & 0.0              \\
\textbf{temp}     & -9.7535                                                               & 0.0              & \textbf{clamt}    & -32.697                                                               & 0.0              \\ \hline
\end{tabular}
}
\caption{Augmented Dickey-Fuller (ADF) Unit Root Test}
\label{tab: adftest}
\end{table}

\subsection{Granger Causality test}
Null hypothesis: The coefficients of past values in the regression equation is zero.
\\\\Before building the model, Granger's causality F test for all possible combinations of time series was implemented to test the causal relationship among them.
\\\\Looking at the results, although rain is not useful information for predicting other series, the other variables in the system are interchangeably causing each other. Therefore, this system of multi time series is considered a good candidate for using the VAR model to forecast.

\subsection{Training models}

\subsubsection{Training VAR}
We respectively iterate the model for different combinations of parameters including p (lag order) and trend.  As can be seen from Table \ref{tab: gridsearchresult}, with the lowest RMSE score of 745.477393, lag order of 36 along with constant trend ‘ct’ is the optimal combination. We then fit the model again with these two parameters.

\begin{table}[!h]
\centering
\resizebox{\columnwidth}{!}{%
\begin{tabular}{|c|c|c|c|c|}
\hline
\textbf{Lag order/Trend} & \textbf{c} & \textbf{ct}                   & \textbf{ctt} & \textbf{n} \\ \hline
\textbf{12}              & 1663.05    & 1481.21                       & 1478.71      & 1661.64    \\ \hline
\textbf{24}              & 917.13     & 876.58                        & 874.93       & 913.19     \\ \hline
\textbf{36}              & 773.62     & {\color[HTML]{FF0000} 745.47} & 754.38       & 769.90     \\ \hline
\textbf{48}              & 935.54     & 896.88                        & 907.27       & 929.04     \\ \hline
\end{tabular}%
}
\caption{Grid Search results}
\label{tab: gridsearchresult}
\end{table}

\subsubsection{Testing for Residuals}
We use Durbin Watson Statistics to check for serial correlation of residuals. The figures are all very close to 2.0, which implies there is no significant serial correlation. Therefore, we can assume that this model is sufficiently able to explain the variances and patterns in the time series.

\begin{table}[!h]
\centering
\resizebox{\columnwidth}{!}{%
\begin{tabular}{
>{\columncolor[HTML]{F3F3F3}}c c
>{\columncolor[HTML]{F3F3F3}}c c}
\hline
\textbf{Variable} & \textbf{\begin{tabular}[c]{@{}c@{}}Durbin Watson\\ Statistic\end{tabular}} & \textbf{Variable} & \textbf{\begin{tabular}[c]{@{}c@{}}Durbin Watson\\ Statistic\end{tabular}} \\ \hline
\textbf{station0} & 2.01                                                                       & \textbf{wetb}     & 2.0                                                                        \\
\textbf{station1} & 2.02                                                                       & \textbf{dewpt}    & 2.0                                                                        \\
\textbf{station2} & 2.03                                                                       & \textbf{rhum}     & 2.0                                                                        \\
\textbf{station3} & 2.02                                                                       & \textbf{vappr}    & 2.0                                                                        \\
\textbf{station4} & 2.01                                                                       & \textbf{msl}      & 2.0                                                                        \\
\textbf{station5} & 2.01                                                                       & \textbf{wdsp}     & 2.0                                                                        \\
\textbf{station6} & 2.02                                                                       & \textbf{wddir}    & 2.0                                                                        \\
\textbf{station7} & 2.02                                                                       & \textbf{sun}      & 2.0                                                                        \\
\textbf{station8} & 2.02                                                                       & \textbf{vis}      & 2.0                                                                        \\
\textbf{rain}     & 2.0                                                                        & \textbf{clht}     & 2.0                                                                        \\
\textbf{temp}     & 2.0                                                                        & \textbf{clamt}    & 2.0                                                                        \\ \hline
\end{tabular}%
}
\caption{Durbin Watson Statistic of Residuals}
\label{durinwatsonstatistic}
\end{table}

Hypothesis of normality test:

H0 : Data generated by normally-distributed process

H1 : Data are not normally distributed
\\\\According to Table \ref{tab:jarqueberatest}, since the Jarque-Bera Statistic is greater than the critical value with p-value less than 0.05, null hypothesis can be rejected at 5\% significance level. Thus, we can conclude that the errors are not normally distributed. We also check the stability of residuals by calculating the root of the characteristic polynomial. The result shows that the modulus of all roots are greater than 1. Therefore, the errors are not stable and the model VAR is said to have low stability.
\begin{table}[!h]
\centering
\begin{tabular}{ccc}
\hline
\textbf{Test Statistic} & \textbf{Critical value} & \textbf{P value} \\ \hline
1.899e+08              & 60.48                   & 0.000            \\ \hline
\end{tabular}%
\caption{The result of Jarque-Bera test for normality}
\label{tab:jarqueberatest}
\end{table}
\\The white noise component in the VAR process is assumed to be normally distributed. While this assumption is not required for parameter estimates to be consistent or asymptotically normal, results are generally more reliable in finite samples when residuals are Gaussian white noise. Our team has tried several approaches to make the residuals stable and have a normal distribution. However, due to limited knowledge about this field, we still have not succeeded. Therefore, we decided to take the risk of unreliable model assumptions and continue on to the next step.

\subsection{Performance Analysis}
Taking data from 01/03/2013 to 27/02/2019 as forecast input, we predict the hourly traffic volume in 28/02/2019. The evaluation scores of the prediction are shown in Table \ref{tab: evalutionmetricsforpredict}, including mean absolute error score (MAE), mean squared error score (MSE) and root mean squared error score (RMSE). As can be seen from Table \ref{tab: evalutionmetricsforpredict}, VAR(36) makes a good prediction on station0, station4, station5 and station6 while the others receive relatively high error scores. This result implies the non-stability of model VAR(36). 
\begin{table}[!h]
\centering
\resizebox{\columnwidth}{!}{%
\begin{tabular}{c|ccc|}
\cline{2-4}
                                        & \multicolumn{3}{c|}{\textbf{Evaluation metrics}}                                      \\ \cline{2-4} 
                                        & \multicolumn{1}{c|}{\textbf{MAE}} & \multicolumn{1}{c|}{\textbf{MSE}} & \textbf{RMSE} \\ \hline
\multicolumn{1}{|c|}{\textbf{station0}} & \multicolumn{1}{c|}{329.2225}     & \multicolumn{1}{c|}{140050.9629}  & 374.2338      \\ \hline
\multicolumn{1}{|c|}{\textbf{station1}} & \multicolumn{1}{c|}{636.4661}     & \multicolumn{1}{c|}{799515.7512}  & 894.1564      \\ \hline
\multicolumn{1}{|c|}{\textbf{station2}} & \multicolumn{1}{c|}{543.3932}     & \multicolumn{1}{c|}{625428.1601}  & 790.8402      \\ \hline
\multicolumn{1}{|c|}{\textbf{station3}} & \multicolumn{1}{c|}{401.0084}     & \multicolumn{1}{c|}{276376.5316}  & 525.7153      \\ \hline
\multicolumn{1}{|c|}{\textbf{station4}} & \multicolumn{1}{c|}{97.3009}      & \multicolumn{1}{c|}{14847.9573}   & 121.8522      \\ \hline
\multicolumn{1}{|c|}{\textbf{station5}} & \multicolumn{1}{c|}{145.3302}     & \multicolumn{1}{c|}{45273.892}    & 212.7766      \\ \hline
\multicolumn{1}{|c|}{\textbf{station6}} & \multicolumn{1}{c|}{168.2309}     & \multicolumn{1}{c|}{46914.3516}   & 216.5972      \\ \hline
\multicolumn{1}{|c|}{\textbf{station7}} & \multicolumn{1}{c|}{637.6432}     & \multicolumn{1}{c|}{907202.9476}  & 952.472       \\ \hline
\multicolumn{1}{|c|}{\textbf{station8}} & \multicolumn{1}{c|}{664.2259}     & \multicolumn{1}{c|}{994103.9514}  & 997.0476      \\ \hline
\multicolumn{1}{|l|}{\textbf{Average}}  & \multicolumn{1}{l|}{402.5357}  & \multicolumn{1}{l|}{427746.0562}  & \multicolumn{1}{l|}{565.0768111} \\ \hline
\end{tabular}%
}
\caption{MAE, MSE and RMSE score for the prediction of each variable and the average score of all variables}
\label{tab: evalutionmetricsforpredict}
\end{table}

\begin{figure} 
    \centering
  \subfloat[Result at station 4\label{station4}]{%
       \includegraphics[width=8cm]{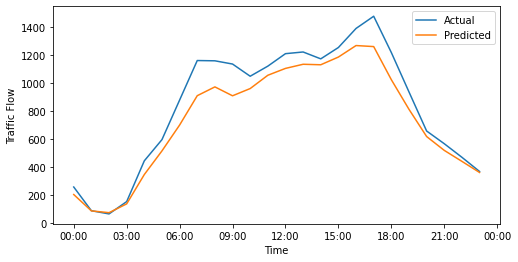}}
    \hfill
  \subfloat[Result at station 8\label{station8}]{%
        \includegraphics[width=8cm]{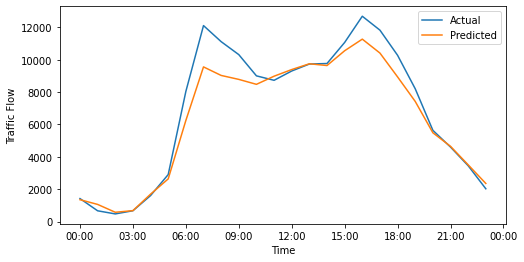}}
    
  \caption{Forecast of traffic flow on 28 February 2019}
  \label{fig:resultpredict} 
\end{figure}

To understand the errors more briefly, the forecast of traffic flow at station4 and station8 are plotted as in Figure \ref{fig:resultpredict}, which are respectively the best and worst prediction. Generally, the forecast looks quite impressive as the model can catch the trend of traffic flow. However, the performance in peak hours, in particular, in 7 a.m, 4 p.m and 5 p.m, is not that accurate.

\subsection{Variance Decomposition}
In order to determine the influential level of each variable on traffic flow at each station, variance decomposition with the lag order of 36 is implemented. Based on the decomposition result, weather variables generally play a less important role in forecasting compared to traffic data. Traffic flow at station 0 strongly predicts itself.  4.17\% of the change in traffic flow at station 0 is explained by traffic flow at station 6. Station 7 and 8, which explain only about 0.3\% of traffic flow, do not have any significant influence on station 0. In the first hour, station 1 is a good predictor of itself, accounts for 71.9\%. This figure, however, reduces to 36.3\% in the next 36 hours. Station 0 explains 38.6\% of the variation in traffic volume at station 1, which makes station 0 the strongest predictor in the long run. Station 7 and 8 do not significantly influence station 1.

The interpretations for the other stations are similar. They weakly influence themselves in both short run and long run. Station 0 and 1 are always the most significant variables, while station 7 and 8 often have no impact on the change of traffic volume.

Among weather attributes, air temperature, wet bulb temperature, vapour pressure, mean sea level pressure and sunshine duration are the most significant predictors, though their influences on traffic flow at each station are not strong enough.

\section{Conclusion and Future Research}
This paper proposes a multivariate approach for traffic volume prediction in one day, having the data of the previous 6 years. Model VAR(36) is used for the forecast in spite of its unreliability. This model produces the performance of 565.0768111 RMSE on average. This promising result indicates that VAR(36) with a time trend and constant is an appropriate model for multi time series forecasting traffic flow in 24 hours. Several methods were also conducted to determine the significance of each variable in forecasting. Traffic data, in general, are more useful than weather information. Data of station 7 and 8 do not show much influence on the traffic flow at the given road segments. Furthermore, rain, dew point temperature, relative humidity, mean wind speed, predominant wind direction, visibility, cloud height and cloud amount are proved to be unnecessary components. 

Our study still contains some limitations such as an unstable and non-normality model. In future research, our team will continue studying methods to make the residuals stable and normally distributed, thus, ensure the reliability of model VAR(36). We also intend to withdraw unnecessary variables to check whether this method will improve the performance of the VAR(36). Furthermore, we will try more complicated time series models such as VARMA, ARIMA, SARIMA,... and combine with several deep learning techniques to solve the posed problem more effectively.

\section*{Acknowledgment} This research was funded by University of Information Technology - Vietnam National University HoChiMinh City under grant number D1-2022-48.

\nocite{*} 
\bibliographystyle{IEEEtran}
\bibliography{bibliography}

\begin{thebibliography}{10}
\providecommand{\url}[1]{#1}
\csname url@samestyle\endcsname
\providecommand{\newblock}{\relax}
\providecommand{\bibinfo}[2]{#2}
\providecommand{\BIBentrySTDinterwordspacing}{\spaceskip=0pt\relax}
\providecommand{\BIBentryALTinterwordstretchfactor}{4}
\providecommand{\BIBentryALTinterwordspacing}{\spaceskip=\fontdimen2\font plus
\BIBentryALTinterwordstretchfactor\fontdimen3\font minus \fontdimen4\font\relax}
\providecommand{\BIBforeignlanguage}[2]{{%
\expandafter\ifx\csname l@#1\endcsname\relax
\typeout{** WARNING: IEEEtran.bst: No hyphenation pattern has been}%
\typeout{** loaded for the language `#1'. Using the pattern for}%
\typeout{** the default language instead.}%
\else
\language=\csname l@#1\endcsname
\fi
#2}}
\providecommand{\BIBdecl}{\relax}
\BIBdecl

\bibitem{dissanayake_2021_4514955}
\BIBentryALTinterwordspacing
B.~Dissanayake, O.~Hemachandra, N.~Lakshitha, D.~Haputhanthri, and A.~Wijayasiri, ``A comparison of arimax, var and lstm on multivariate short-term traffic volume forecasting,'' Feb 2021. [Online]. Available: \url{https://doi.org/10.5281/zenodo.4514955}
\BIBentrySTDinterwordspacing

\bibitem{10.1063/1.5016666}
\BIBentryALTinterwordspacing
A.~Suharsono, A.~Aziza, and W.~Pramesti, ``{Comparison of vector autoregressive (VAR) and vector error correction models (VECM) for index of ASEAN stock price},'' \emph{AIP Conference Proceedings}, vol. 1913, no.~1, p. 020032, 12 2017. [Online]. Available: \url{https://doi.org/10.1063/1.5016666}
\BIBentrySTDinterwordspacing

\bibitem{Zakai2014ATS}
\BIBentryALTinterwordspacing
M.~Zakai, ``A time series modeling on gdp of pakistan,'' 2014. [Online]. Available: \url{https://api.semanticscholar.org/CorpusID:55818597}
\BIBentrySTDinterwordspacing

\bibitem{gmd-7-1247-2014}
\BIBentryALTinterwordspacing
T.~Chai and R.~R. Draxler, ``Root mean square error (rmse) or mean absolute error (mae)? – arguments against avoiding rmse in the literature,'' \emph{Geoscientific Model Development}, vol.~7, no.~3, pp. 1247--1250, 2014. [Online]. Available: \url{https://gmd.copernicus.org/articles/7/1247/2014/}
\BIBentrySTDinterwordspacing

\bibitem{Iwok-2016}
I.~Iwok and A.~Okpe, ``A comparative study between univariate and multivariate linear stationary time series models,'' \emph{American Journal of Mathematics and Statistics}, vol. 2016, pp. 203--212, 07 2016.

\bibitem{anile-2007}
A.~Seth, ``Granger causality,'' \emph{Scholarpedia}, vol.~7, p. 1667, 07 2007.

\bibitem{pedro-2006}
P.~Galeano, D.~Peña, and R.~Tsay, ``Outlier detection in multivariate time series by projection pursuit,'' \emph{Journal of the American Statistical Association}, vol. 101, pp. 654--669, 06 2006.

\bibitem{cheung-1995}
\BIBentryALTinterwordspacing
Y.-W. Cheung and K.~S. Lai, ``Lag order and critical values of the augmented dickey–fuller test,'' \emph{Journal of Business \& Economic Statistics}, vol.~13, no.~3, pp. 277--280, 1995. [Online]. Available: \url{https://doi.org/10.1080/07350015.1995.10524601}
\BIBentrySTDinterwordspacing

\bibitem{usman-2017}
M.~Usman, D.~F. Fatin, M.~Y.~S. Barusman, F.~A.~M. Elfaki, and Widiarti, ``Application of vector error correction model (vecm) and impulse response function for analysis data index of farmers’ terms of trade,'' \emph{Indian Journal of Science and Technology}, vol.~10, pp. 1--14, 02 2017.

\bibitem{Hong2010}
\BIBentryALTinterwordspacing
Y.~Hong, \emph{Serial correlation and serial dependence}.\hskip 1em plus 0.5em minus 0.4em\relax London: Palgrave Macmillan UK, 2010, pp. 227--244. [Online]. Available: \url{https://doi.org/10.1057/9780230280830_25}
\BIBentrySTDinterwordspacing

\bibitem{helmut-2015}
H.~Luetkepohl, \emph{The New Introduction to Multiple Time Series Analysis}, 01 2005.

\end{thebibliography}

\end{document}